\documentclass{article}

\usepackage{times}
\usepackage{graphicx} % more modern
\usepackage{subfigure} 

\usepackage{natbib}

\usepackage{amsmath,amssymb}

\usepackage{algorithm}
\usepackage{algorithmic}

\usepackage{hyperref}

\newcommand{\vect}[1]{\mathbf{#1}}
\newcommand{\vects}[1]{\boldsymbol{#1}}
\newcommand{\matr}[1]{\mathbf{#1}}
\newcommand{\vh}[0]{\vect{h}}
\newcommand{\vx}[0]{\vect{x}}
\newcommand{\TT}[0]{{\vects{\theta}}}
\newcommand{\vb}[0]{\vect{b}}
\newcommand{\vc}[0]{\vect{c}}
\newcommand{\mW}[0]{\matr{W}}

\usepackage[accepted]{icml2014} 
\icmltitlerunning{Gradient Estimate Variance in CD and PCD}

\begin{document} 

\twocolumn[
\icmltitle{Stochastic Gradient Estimate Variance in Contrastive Divergence  \\ 
           and Persistent Contrastive Divergence}

% unless you've provided the [accepted] option to the icml2014
\icmlauthor{Mathias Berglund}{mathias.berglund@aalto.fi}
\icmladdress{Aalto University,
            Finland}
\icmlauthor{Tapani Raiko}{tapani.raiko@aalto.fi}
\icmladdress{Aalto University,
            Finland}

\icmlkeywords{Restricted Boltzmann Machines, Contrastive Divergence, Persistent Contrastive Divergence}

\vskip 0.3in
]

\begin{abstract} 
Contrastive Divergence (CD) and Persistent Contrastive Divergence (PCD) are popular methods for training the weights of Restricted Boltzmann Machines. However, both methods use an approximate method for sampling from the model distribution. As a side effect, these approximations yield significantly different biases and variances for stochastic gradient estimates of individual data points. It is well known that CD yields a biased gradient estimate. In this paper we however show empirically that CD has a lower stochastic gradient estimate variance than exact sampling, while the mean of subsequent PCD estimates has a higher variance than exact sampling. The results give one explanation to the finding that CD can be used with smaller minibatches or higher learning rates than PCD.
\end{abstract} 

\section{Introduction}
Popular methods to train Restricted Boltzmann Machines \cite{smolensky1986information} include Contrastive Divergence \cite{hinton2002training, hinton2006reducing} and Persistent Contrastive Divergence\footnote{PCD is also known as Stochastic Maximum Likelihood} \cite{younes1989parametric, tieleman2008training}. Although some theoretical research has focused on the properties of these two methods \cite{bengio2009justifying, carreira2005contrastive, tieleman2008training}, both methods are still used in similar situations, where the choice is often based on intuition or heuristics.

One known feature of Contrastive Divergence (CD) learning is that it yields a biased estimate of the gradient \cite{bengio2009justifying, carreira2005contrastive}. On the other hand, it is known to be fast for reaching good results \cite{carreira2005contrastive, tieleman2008training}. In addition to the computationally light sampling procedure in CD, it is claimed to benefit from a low variance of the gradient estimates \cite{hinton2002training, carreira2005contrastive}. However, the current authors are not aware of any rigorous research on whether this claim holds true, and what the magnitude of the effect is\footnote{The topic has been covered in e.g. \cite{williams2002analysis}, although for a Boltzmann machine with only one visible and hidden neuron.}.

On the other hand, Persistent Contrastive Divergence (PCD) has empirically been shown to require a lower learning rate and longer training than CD\footnote{There are however tricks to be able to increase the learning rate of PCD, see e.g. \cite{swersky2010tutorial}} \cite{tieleman2008training}. The authors propose that the low learning rate is required since the model weights are updated while the Markov chain runs, which means that in order to sample from a distribution close to the stationary distribution the weight cannot change too rapidly. However, for similar reasons that CD updates are assumed to have low variance, subsequent PCD updates are likely to be correlated leading to a possibly undesirable "momentum" in the updates. This behavior would effectively increase the variance of the mean of subsequent updates, requiring either larger minibatches or smaller learning rates.

In this paper we explore the variances of CD, PCD and exact stochastic gradient estimates. By doing so, we hope to shed light on the observed fast speed of CD learning, and on the required low learning rate for PCD learning compared to CD learning. Thereby we hope to contribute to the understanding of the difference between CD and PCD beyond the already well documented bias of CD.

\section{Contrastive Divergence and Persistent Contrastive Divergence}

A restricted Boltzmann machine (RBM) is a Boltzmann machine where each visible neuron $x_i$ is connected to all hidden neurons $h_j$ and each hidden neuron to all visible neurons, but there are no edges between the same type of neurons. An RBM defines an energy of each state $\left(\vx, \vh\right)$ by $-E(\vx, \vh \mid \TT) = \vb^\top \vx + \vc^\top \vh + \vx^\top \mW \vh$, and assigns the following probability to the state via the Boltzmann distribution: 
$
p(\vx, \vh \mid \TT) = \frac{1}{Z(\TT)} \exp \left\{-E\left(\vx , \vh \mid \TT\right)\right\},
$
where $\TT=\left\{ \vb, \vc, \mW \right\}$ is a set of parameters and $Z(\TT)$ normalizes the probabilities to sum up to one.

The log likelihood of one training data point is hence $\phi = \log P(\vx) = \log \left( \sum_{\vh} \exp \left\{-E\left(\vx , \vh \mid \TT\right)\right\} \right) - \log Z(\TT) = \phi^+ - \phi^-$. Sampling the \emph{positive phase} of the gradient of the log likelihood $\frac{\partial \phi^+}{\partial \mW}$ is easy, but sampling the \emph{negative phase} $\frac{\partial \phi^-}{\partial \mW}$ is intractable.

A popular method to solve sampling of the negative phase is Contrastive Divergence (CD). In CD, the negative particle is sampled only approximately by running a Markov Chain a limited number of steps (often only one step) from the positive particle \cite{hinton2002training}. Another method, called Persistent Contrastive Divergence (PCD) solves the sampling with a related method, only that the negative particle is not sampled from the positive particle, but rather from the negative particle from the last data point \cite{tieleman2008training}.

\section{Experiments}

In order to examine the variance of CD and PCD gradient estimates, we use an empirical approach. We train an RBM and evaluate the variance of gradient estimates from different sampling strategies at different stages of the training process. The sampling strategies are CD-k with k ranging from 1 to 10, PCD, and CD-1000 that is assumed to correspond to an almost unbiased stochastic gradient. In addition, we test CD-k with independent samples (I-CD), where the negative particle is sampled from a random training example. The variance of I-CD separates the effect of the negative particle being close to the data distribution in general, and the effect of the negative particle being close to the positive particle in question.

We use three different data sets. The first is a reduced size MNIST \cite{lecun1998gradient} set with $14 \times 14$ pixel images of the first 1~000 training set data points of each digit, totaling 10~000 data points. The second data set are the center $14 \times 14$ pixels of the first 10~000 CIFAR \cite{krizhevsky2009learning} images converted into gray scale. The third are the Caltech 101 Silhouettes \cite{marlin2010inductive}, with 8~641 $16 \times 16$ pixel black and white images. We binarize the grayscale images by sampling the visible units with activation probabilities equal to the pixel intensity.

We set the number of hidden neurons equal to the number of visible neurons. The biases are initialized to zero, while the weights are initially sampled from a zero-mean normal distribution with standard deviation $1/\sqrt{n_{v} + n_{h}}$ where $n_{v}$ and $n_{h}$ are the number of visible and hidden neurons, respectively. We train the model with CD-1, and evaluate the variance of the gradient estimates after 10, and 500 epochs. We use Adaptive learning rate \cite{cho2011enhanced} with an initial learning rate of 0.01. We do not use weight decay.

In all of the gradient estimates, the final sampling step for the probabilities of the hidden unit activations is omitted. The gradient estimate is therefore based on sampled binary visible unit activations, but continuous hidden unit activation probabilities conditional on these visible unit activations. This process is called Rao-Blackwellisation \cite{swersky2010tutorial}, and is often used in practice. The variance is calculated on individual gradient estimates based on only one positive and negative particle each. In practice, the gradient is usually estimated by averaging over a mini-batch of N independent samples, which diminishes the variance N-fold. We ignore the bias gradient estimates.

%\begin{figure}[t]
%\vskip 0.2in
%\begin{center}
%\centerline{\includegraphics[width=\columnwidth]{figs/cd_var_early.pdf}}
%\caption{CD-k gradient estimate variance in large system for different values of k compared to CD-1000. Error bars indicate standard deviation between iterations.}
%\label{cd_var_early}
%\end{center}
%\vskip -0.2in
%\end{figure} 

\begin{figure}[t]
\vskip 0.2in
\begin{center}
\centerline{\includegraphics[width=\columnwidth]{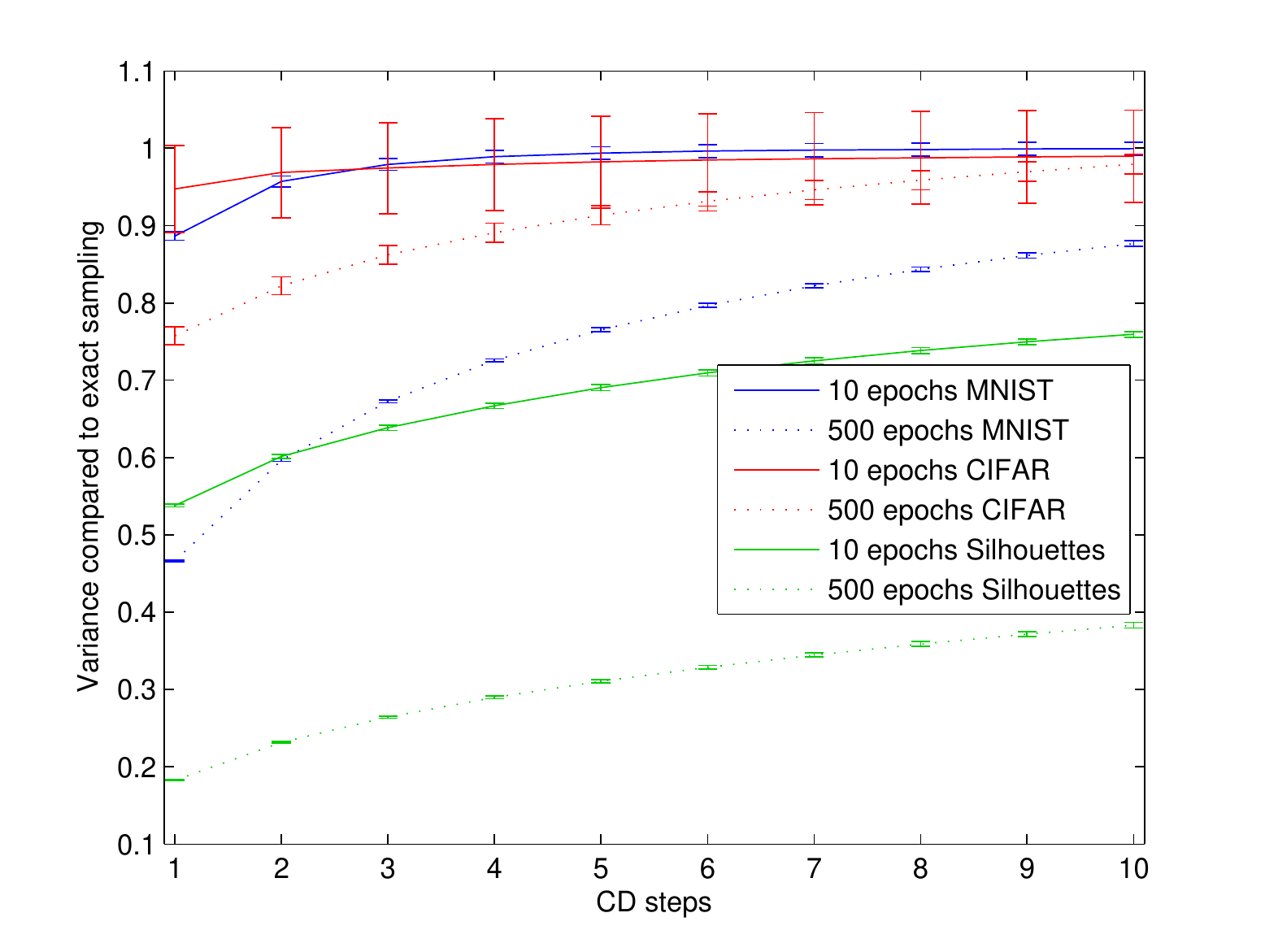}}
\caption{CD-k gradient estimate variance for different values of k compared to CD-1000 after 10 and 500 epochs of training. Error bars indicate standard deviation between iterations.}
\label{cd_var_late}
\end{center}
\vskip -0.2in
\end{figure}

When analyzing subsequent PCD gradient estimates, the negative particles of the first estimate are sampled 1~000 steps from a random training example. Subsequent k estimates are then averaged, where the positive particle is randomly sampled from the data for each step while the negative particle is sampled from the previous negative particle. No learning occurs between the subsequent estimates. We can therefore disentangle the effects of weight updates from the effect of correlation between subsequent estimates.

We iterate all results for 10 different random initializations of the weights, and evaluate the variance by sampling gradient estimates of individual training examples 10 times for each training example in the data set. The variance is calculated for each weight matrix element separately, and the variances of the individual weights are then averaged.

\section{Results}

As we can see from Figure~\ref{cd_var_late}, the variance of Contrastive Divergence is indeed smaller than for exact sampling of the negative particle. We also see that the variance of CD estimates quickly increases with the number of CD steps. However, this effect is significant only in later stages of training. This phenomenon is expected, as the model is expected not to mix as well in later stages of training as when the weights are close to the small initial random weights.

\begin{figure}[t]
\vskip 0.2in
\begin{center}
\centerline{\includegraphics[width=\columnwidth]{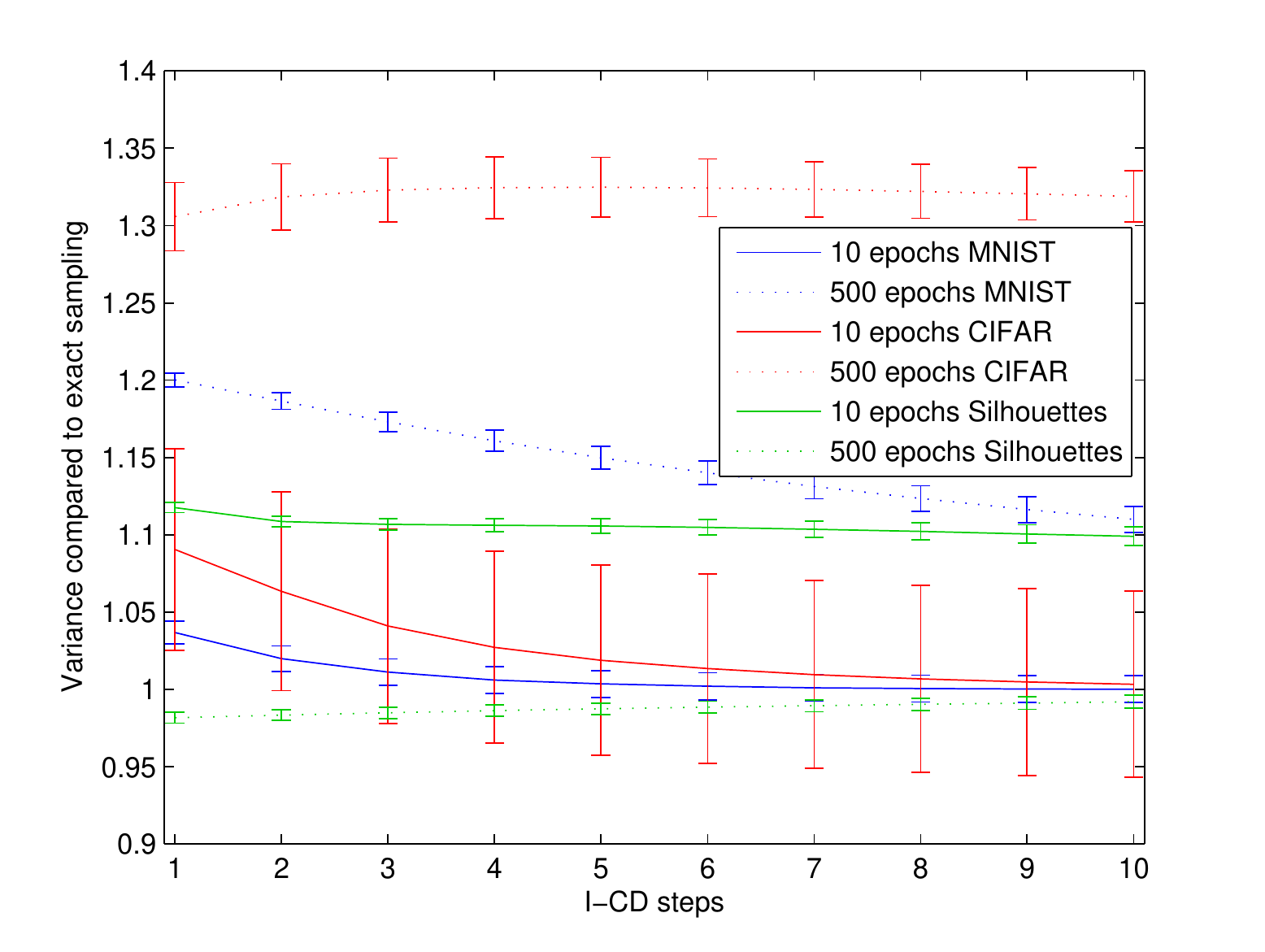}}
\caption{I-CD-k gradient estimate variance for different values of k compared to CD-1000 after 10 and 500 epochs of training. Error bars indicate standard deviation between iterations.}
\label{cd_faux_var_late}
\end{center}
\vskip -0.2in
\end{figure}

If we sample the negative particle from a different training example than the positive particle (I-CD), in Figure~\ref{cd_faux_var_late} we see that the variance is similar or even larger compared to the variance with exact sampling. Although it is trivial that the variance of the I-CD estimates is higher than for CD, the interesting result is that I-CD loses all of the variance advantage against exact sampling. The result supports the hypothesis that the low variance of CD precisely stems from the fact that the negative particle is sampled from the positive particle, and not from that the negative particle is sampled only a limited number of steps from a random training example.

\begin{figure}[t]
\vskip 0.2in
\begin{center}
\centerline{\includegraphics[width=\columnwidth]{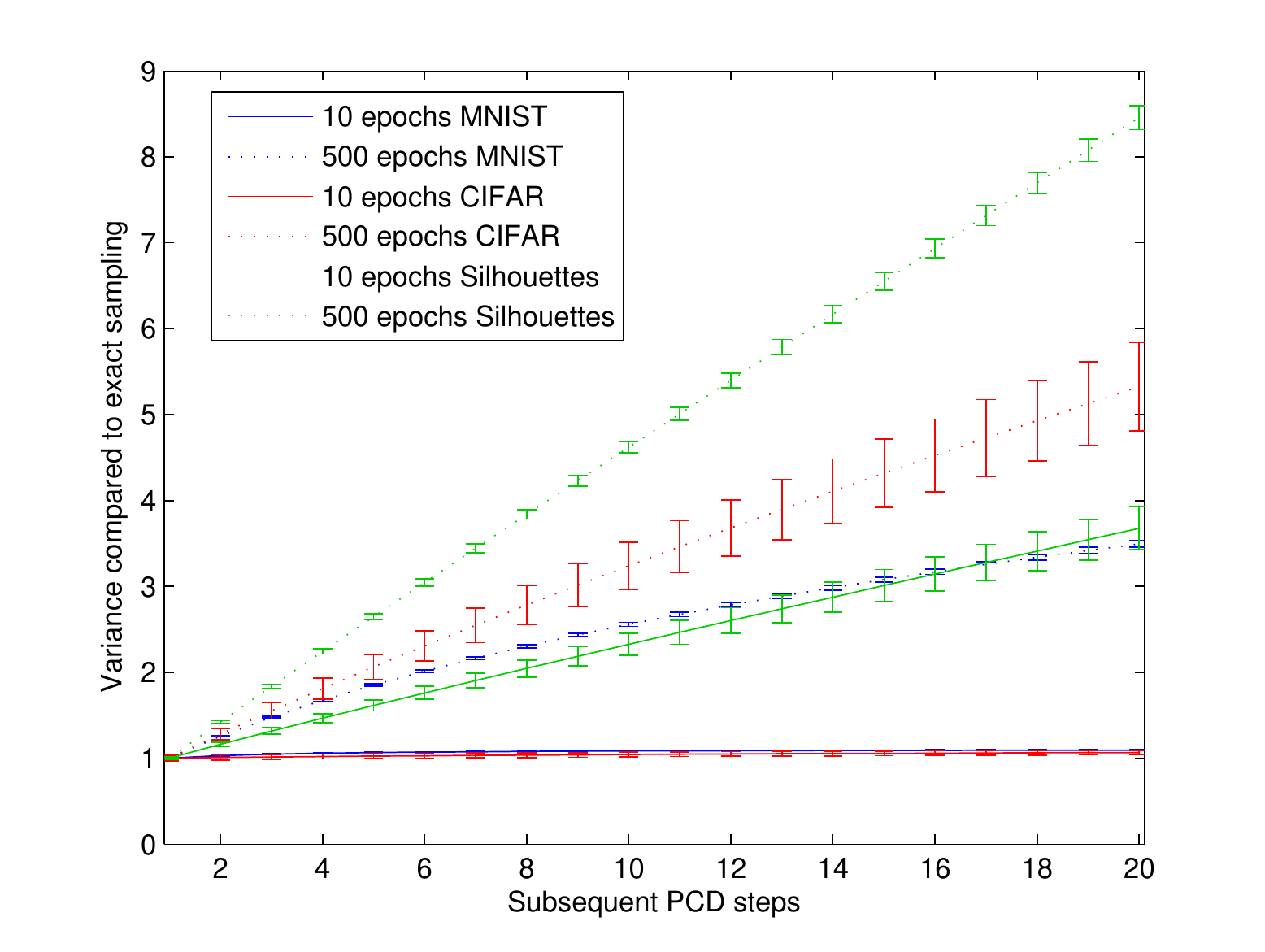}}
\caption{The PCD vs CD-1000 ratio of the variance for the mean of k subsequent estimates after 10 and 500 epochs of training. Error bars indicate standard deviation between iterations.}
\label{pcd_var_late}
\end{center}
\vskip -0.2in
\end{figure} 

For subsequent PCD updates, we see in Figure~\ref{pcd_var_late} that the variance indeed is considerably higher than for independent sampling. Again, as expected this effect is stronger the later during training the evaluation is done.

%, and in Figure~\ref{pcd_var_early} we see that this effect is practically not observable in the beginning of training.

%\begin{figure}[t]
%\vskip 0.2in
%\begin{center}
%\centerline{\includegraphics[width=\columnwidth]{figs/pcd_var_early.pdf}}
%\caption{PCD gradient estimate variance for sum of k subsequent estimates compared to CD-1000. Error bars indicate standard deviation between iterations.}
%\label{pcd_var_early}
%\end{center}
%\vskip -0.2in
%\end{figure} 

When looking at the magnitude of the variance difference, we see that for CD-1, the mean of 10 subsequent updates have a multiple times smaller variance than PCD. In effect, this means that ignoring any other effects and the effect of weight updates, PCD would need considerably smaller learning rates or larger minibatches to reach the same variance per minibatch. This magnitude is substantial, and might explain the empirical finding that PCD performs best with smaller learning rates than CD.

\section{Conclusions}

Contrastive Divergence or Persistent Contrastive Divergence are often used for training the weights of Restricted Boltzmann machines. Contrastive Divergence is claimed to benefit from low variance of the gradient estimates when using stochastic gradients. Persistent Contrastive Divergence could on the other hand suffer from high correlation between subsequent gradient estimates due to poor mixing of the Markov chain estimating the model distribution.

In this paper, we have empirically confirmed both of these findings. In experiments on three data sets, we find that the variance of CD-1 gradient estimates are considerably lower than when independently sampling with many steps from the model distribution. Conversely, the variance of the mean of subsequent gradient estimates using PCD is significantly higher than with independent sampling. This effect is mainly observable towards the end of training. In effect, this indicates that from a variance perspective, PCD would require considerably lower learning rates or larger minibatches than CD. As CD is known to be a biased estimator, it therefore seems that the choice between CD and PCD is a trade-off between bias and variance.

%The merits of CD therefore seem to lie somewhere else than in low variance of the gradient estimates. In addition, PCD does not seem to be hampered by the correlation of subsequent gradient estimates, at least from the perspective of this introducing unwanted "momentum" in subsequent gradient estimates.

Although the results in this paper are practically significant, the approach in this paper is purely empirical. Further theoretical analysis of the variance of PCD and CD gradient estimates would therefore be warranted to confirm these findings. In addition, we intend to repeat the experiments with a larger data set, and train the models and replace the CD-1000 baseline with better approximations of exact sampling using e.g. Enhanced gradient \cite{cho2011enhanced} and parallel tempering \cite{cho2010parallel}.

\bibliography{bibliography}
\bibliographystyle{icml2014}

\end{document}